\documentclass[11pt,a4paper]{article}
\usepackage[hyperref]{acl2019}
\usepackage{times}
\usepackage{latexsym}

\usepackage{stackrel,amssymb}
\usepackage{url}
\usepackage{color,soul}
\usepackage[normalem]{ulem}
\usepackage{textcomp}
\usepackage{todonotes}

\usepackage{booktabs}
\usepackage{multirow}
\usepackage{comment}
\usepackage{siunitx}

\usepackage{CJKutf8}
\usepackage{textcomp}
\newcommand{\emoji}[1]{\includegraphics[width=1em]{emoji_images/#1.png}}

\aclfinalcopy 

\newcommand*{\metric}[1]{\num[round-mode=places,round-precision=1]{#1}}

\title{Findings of the First Shared Task on \\ Machine Translation Robustness}

\author{
  Xian Li$^1$, Paul Michel$^2$,  Antonios Anastasopoulos$^2$, Yonatan Belinkov$^3$, Nadir Durrani$^4$,\\
  \textbf{Orhan Firat$^5$, Philipp Koehn$^6$, Graham Neubig$^2$, Juan Pino$^1$, Hassan Sajjad$^4$} \\
  $^1$Facebook AI, 
$^2$Carnegie Mellon University, $^3$Harvard University and MIT, \\ 
$^4$Qatar Computing Research Institute, $^5$Google AI, $^6$Johns Hopkins University\\
 }
\date{}

\begin{document}

\maketitle
\begin{abstract}
  We share the findings of the first shared task on improving robustness of Machine Translation (MT). The task provides a testbed representing challenges facing MT models deployed in the real world, and facilitates new approaches to improve models' robustness to noisy input and domain mismatch. We focus on two language pairs (English-French and English-Japanese), and the submitted systems are evaluated on a blind test set consisting of noisy comments on Reddit\footnote{\url{www.reddit.com}} and professionally sourced translations. As a new task, we received 23 submissions by 11 participating teams from universities, companies, national labs, etc. All submitted systems achieved large improvements over baselines, with the best improvement having $+22.33$ BLEU. 
  We evaluated submissions by both human judgment and automatic evaluation (BLEU), which shows high correlations (Pearson's $r=0.94$ and $0.95$). Furthermore, we conducted a qualitative analysis of the submitted systems using \texttt{compare-mt}\footnote{\url{https://github.com/neulab/compare-mt}}, which revealed their salient differences in handling challenges in this task.
  Such analysis provides additional insights when there is occasional disagreement between human judgment and BLEU, e.g. systems better at producing colloquial expressions received higher score from human judgment.
\end{abstract}

\section{Introduction}
In recent years, Machine Translation (MT) systems have seen great progress, with neural models becoming the \textit{de-facto} methods and  even approaching human quality in news domain \cite{hassan2018achieving}.
However, like other deep learning models, neural machine translation (NMT) models are found to be sensitive to synthetic and natural noise in input, distributional shift, and adversarial examples~\cite{koehn2017six,belinkov2017synthetic,durrani-etal-2019,anastasopoulos+etal:naacl2019,michel2019evaluation}. From an application perspective, MT systems need to deal with non-standard, noisy text of the kind which is ubiquitous on social media and the internet, yet has different distributional signatures from corpora in common benchmark datasets. 

The goal of this shared task is to provide a testbed for improving MT models' robustness to orthographic variations, grammatical errors, and other linguistic phenomena common in user-generated content, via better modelling, training, adaptation techniques, or leveraging monolingual training data.
Specifically, the shared task aims to bring improvements on the following challenges:
\begin{itemize}
    \item To improve NMT's robustness to orthographic variations, grammatical errors, informal language, and other linguistic phenomena or noise common on social media.
    \item To explore effective approaches to leverage abundant out-of-domain parallel data.
    \item To explore novel approaches to leverage abundant monolingual data on the Web (e.g., tweets, Reddit comments, commoncrawl, etc.).
    \item To thoroughly investigate and understand the overall challenges in translating social media text and identify major themes of efforts which needs more research from the community.
\end{itemize}

In this first iteration, the shared-task used the MTNT dataset \citep{michel2018mtnt} that contains noisy social media texts and their translations between English (Eng) and French (Fra) and English and Japanese (Jpn), in four translation directions: Eng$\rightarrow$Fra, Fra$\rightarrow$Eng, Eng$\rightarrow$Jpn, and Jpn$\rightarrow$Eng.  We describe the dataset and the task setup in Section~\ref{sec:task}. The shared-task attracted a total of 23 submissions from 11 teams. The teams employed a variety of methods to improve robustness. A specific challenge was the small size of the in-domain noisy parallel dataset. We summarize  the participating systems in Section~\ref{sec:systems} and the notable methods in Section~\ref{sec:methods}.  The contributions were evaluated both automatically and via a human evaluation. 
The results demonstrate a significant progress of the state-of-the-art in MT robustness, with multiple teams surpassing the shared-task baseline by a large margin. These results are discussed in Section~\ref{sec:results}. 

We hope that this task leads to more efforts from the community in building robust MT models.

\section{Related Work}

The fragility of neural networks \cite{szegedy2013intriguing} has been shown to extend to neural machine translation models \cite{belinkov2017synthetic,heigold2017robust} and recent work focused on various aspects of the problem. From the identification of the causes of this brittleness, to the induction of (adversarial) inputs that trigger the unwanted behavior (attacks) and making such models robust against various types of noisy inputs (defenses); improving robustness has been receiving increasing attention in NMT.

While \citet{koehn2017six} mentioned domain mismatch as a challenge for neural machine translation, \citet{khayrallah-koehn-2018-impact} addressed noisy training data and focus on the types of noise occurring in web-crawled corpora. \citet{michel2018mtnt} proposed a new dataset (MTNT) to test MT models for robustness to the types of noise encountered in the Internet and demonstrated that these challenges cannot be overcome by simple domain adaptation techniques alone. 

\citet{belinkov2017synthetic} and \citet{heigold2017robust} showed that NMT systems are very sensitive to slightly perturbed input forms, and hinted at the importance of injecting noisy examples during training, also known as adversarial examples. Further research proposed several methods of generating and using noisy examples as NMT input to advance the understanding and improve the translation quality. Following machine vision, two major branches being explored when generating noisy examples, \textit{i)} white box methods, where adversarial examples are generated with access to the model parameters \cite{ebrahimi-etal-2018-adversarial,Cheng2018Seq2SickET, cheng-tu-2018-towards, Cheng-etal-2019-doubly} and \textit{ii)} black-box attacks, where examples are generated without accessing model internals \cite{Zhao2018generating,Lee2018HallucinationsIN,Karpukhin2019training,anastasopoulos+etal:naacl2019,Vaibhav-etal-2019-improving}; see~\citet{belinkov2019analysis} for a categorization of such work. 
In particular, some have focused on specific variations of naturally-occurring noise, such as grammatical errors produced by non-native speakers \cite{anastasopoulos+etal:naacl2019} or errors extracted from Wikipedia edits \cite{belinkov2017synthetic}. It has also been shown that adding synthetic noise does not trivially increase robustness to natural noise~\cite{belinkov2017synthetic} and may require specific recipes~\cite{DBLP:journals/corr/abs-1902-01509}. 

\citet{michel2019evaluation} recently emphasized the importance of meaning-preserving perturbations and along with \citet{Cheng-etal-2019-doubly} demonstrated the utility of adversarial training without significantly impairing performance on clean data and domain. \citet{durrani-etal-2019} showed that character-based representations are more robust towards noise compared to such learned using BPE-based sub-word units in the task of machine translation.

\section{Task} \label{sec:task} 

This is the first year we introduce the robustness task. The goal of the task setup is to examine MT systems' performance on non-standard, noisy, user-generated text, which often resemble mixed challenges around orthographic variations, grammar errors, domain shift and stylistic lexical choice, etc. We use the MTNT dataset \cite{michel2018mtnt} as a testbed for the above-mentioned robustness challenges. To give readers an idea of the natural ``noise" present in the MTNT dataset, and the challenges for MT systems to robustly understand and translate them, we provide some examples of input variations:
\begin{itemize}
    \item \textbf{Spelling/typographical errors:} accross (across), recieve (receive), tant (temps)
    \item \textbf{Grammatical errors:} a tons of, there are less people
    \item \textbf{Spoken language and internet slang:} wanna, chais pas, tbh, smh, mdr
    \item \textbf{Code switching:} This is so kawaii, C'est trop mainstream
    \item \textbf{Profanity/slurs:} f*ck, m*rde
\end{itemize}

Readers are encouraged to refer to \newcite{michel2018mtnt} for more details. This year's task probes MT robustness for two language pairs, French to/from English and Japanese to/from English.

\subsection{Task Setup}
The task includes two tracks, \textit{constrained} and \textit{unconstrained} depending on whether the system is trained on a predefined training datasets or not. The two tracks are evaluated by the same automatic and human evaluation protocol, however, they are compared separately.

For the constrained system track, the task specifies two types of training data in addition to MTNT train set:
\begin{itemize}
    \item \textbf{``Out-of-domain" parallel data:} This facilitates MT model's capability to perform supervised learning from examples with different distribution such as lexical choice, language style, genre etc. For example, parallel corpora from WMT news translation task, subtitles and TED talks are specified.
    \item \textbf{Monolingual data:} We encourage participants to develop novel solutions to learn from unlabelled data, improve existing semi-supervised approach such as backtranslation. We provide both in-domain (MTNT) and out-of-domain (News Commentary, News Crawl, etc) monolingual data.
\end{itemize}

\subsection{Training Data} \label{sec:task-data} 
In the constrained setting, participants were allowed to use the WMT15 training data\footnote{\url{http://www.statmt.org/wmt15/translation-task.html}} for Eng$\leftrightarrow$Fra and any of the KFTT \cite{neubig11kftt}, JESC \cite{pryzant_jesc_2017} and TED talks \cite{cettoloEtAl:EAMT2012} corpora for Jpn$\leftrightarrow$Eng. Additionally, the use of the MTNT corpus \cite{michel2018mtnt} was allowed in order to adapt models on limited in-domain data.

\subsection{Test Data} The test sets were collected following the same protocol as the MTNT dataset, \textit{i.e.} collected from Reddit, filtered out for noisy comments using a sub-word language modeling criterion and translated by professional translators. The 
statistics of the test sets are reported in Table \ref{tab:test_sets_stats}.

\begin{table*}[t]
    \centering
    \begin{tabular}{l|cccc}
    \toprule
    & Eng-Fra & Fra-Eng & Eng-Jpn & Jpn-Eng \\
    \midrule
    \# samples & 1,401 & 1,233 & 1,392 & 1,111 \\
    \midrule
    \# source tokens  & 20.0k & 19.8k & 20.0k & 18.7k \\
    \# target tokens & 22.8k & 19.2k & 33.6k& 13.4k\\
    \bottomrule 
    \end{tabular}
    \caption{Statistics of the test sets.}
    \label{tab:test_sets_stats}
\end{table*}

\subsection{Evaluation protocol}
\label{sec:protocol}

\begin{figure*}
\begin{center}
  \includegraphics[scale=0.4]{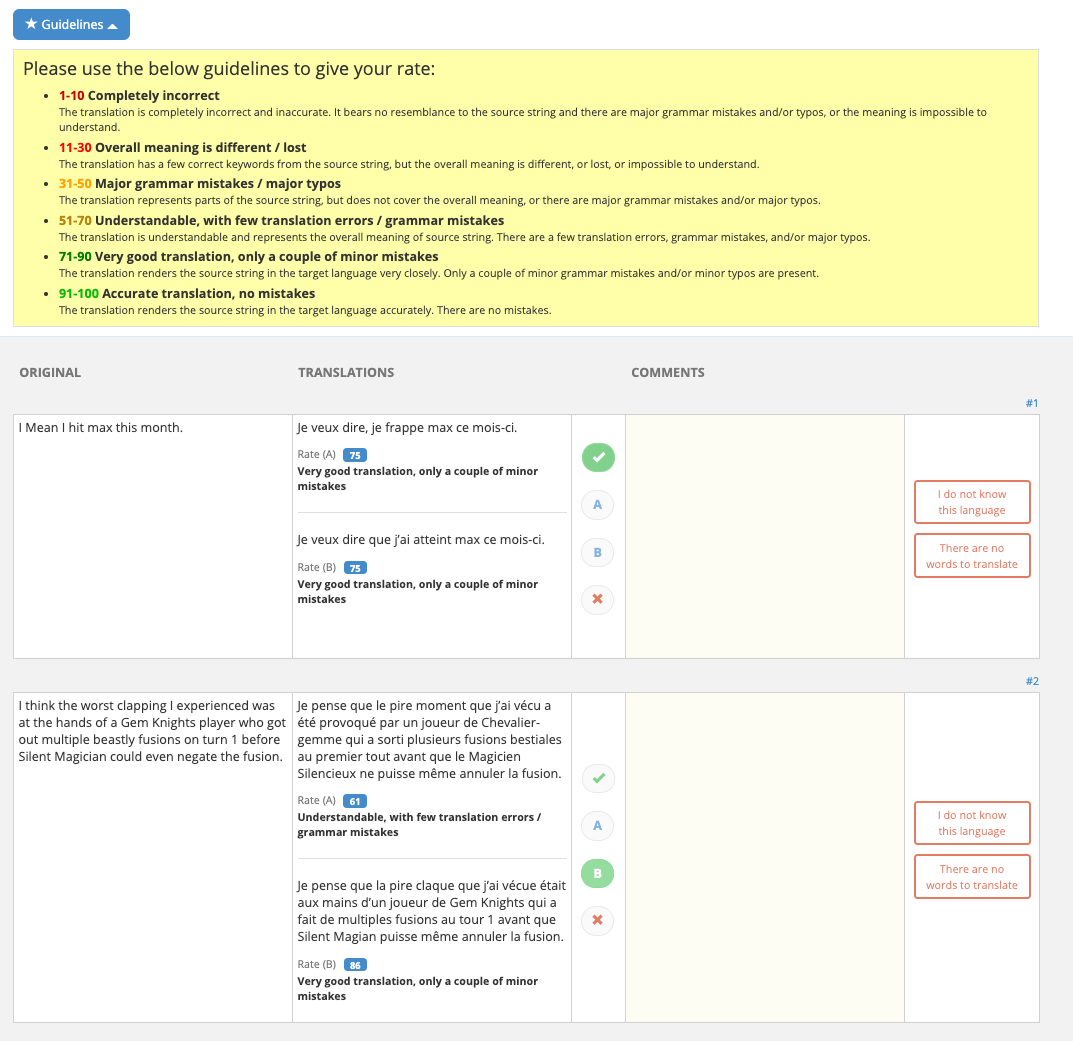}
\caption{Annotation interface for human evaluations.}
\label{fig:annotation_interface}
\end{center}
\end{figure*}

The system outputs were evaluated by professional translators. The translators were presented the original source sentence, the reference and the system output side by side. The order between the reference and the system output was randomized by the user interface. The translators rated both the reference and the translation on a scale from 1 to 100. For both the original source sentence and the reference, the original text was presented except for Eng-Jpn where the Japanese reference tokenized with KyTea was presented in order to be consistent with the systems' outputs. The user interface for annotation is illustrated in Figure \ref{fig:annotation_interface}.

We also evaluated BLEU~\cite{papineni-etal-2002-bleu} for each system using SacreBLEU~\cite{post-2018-call}. For all language pairs except Eng-Jpn, we used the original reference and SacreBLEU with the default options. In the case of Eng-Jpn, we used the reference tokenized with KyTea and the option \texttt{--tokenize none}.

\section{Participants and System Descriptions} \label{sec:systems}

We received 23 submissions from 11 teams. Except two submissions on the Eng-Fra language pair, all systems used the \textit{constrained} setup. Below we briefly describe the systems from the 8 teams which submitted corresponding system description papers:

\paragraph{Baidu \& Oregon State University's submission \cite{baidu}:}
Their system is based on the Transformer implementation in OpenNMT-py~\cite{opennmt}. The main methods applied in their submission are: domain-sensitive data mixing and data augmentation with back-translation. For data mixing, they used a special symbol on the source side to indicate the data domain. 
For data augmentation, they back-translate from a target language to its noisy source. The intuition, also observed by \citet{michel2018mtnt}, is that the source sentences are noisier than their target translations. They include out-of-domain clean data during this step and differentiate data types with a special symbol on the target side. In addition, they also run a model ensemble. 

The team experimented with the Fra$\rightarrow$Eng and Eng$\rightarrow$Fra translation directions, obtaining 43.6 and 36.4 BLEU-cased, respectively (3rd place in both). Their ablations show significant benefit from domain-sensitive training (+3 BLEU), with additional improvements from back-translation and ensembling. 

\paragraph{CMU's submission \cite{cmu}:} This submission only participated in the Fra$\rightarrow$Eng direction. They proposed the use of tied multitask learning, where the noisy source sentences are first decoded by a same-language denoising decoder, and both information is passed on to the translation decoder. This approach requires data triples of noisy source, clean source, translation, which they created by data augmentation over the provided data, using tag-informed translation systems trained on either noisy (MTNT) or clean (Europarl) data. As the participants point out though, their performance improvements seems to be attributed to data augmentation and not to the intermediate denoising decoder.

\paragraph{CUNI's submission \cite{cuni}:}
They participated in Eng$\rightarrow$Fra and Fra$\rightarrow$Eng directions, following a classical two stage approach, i) training of a base model using a mix of parallel (WMT15 Eng-Fra News Translation) and back-translated monolingual data (from News Crawl and Europarl - excluding News Discussions), ii) fine-tuning of the base model using the training portion of the MTNT dataset. All models follow the Transformer-Big architecture, with the hyper-parameters and optimization recipe from the 2018 WMT News Translation shared task submission of CUNI, without ensembles. For both Eng-Fra and Fra-Eng directions, fine-tuning brought about 2+ BLEU points on top of the base models with the Transformer-Big architecture, whereas improvements were substantially larger when the base models were RNN-Based MTNT baselines, about 8+ BLEU points. Participants emphasized the importance of their strong Transformer-Big base model which was already 10+ BLEU points better than the MTNT baseline provided by the shared task. The effect of individual partitions of the base model training set (parallel and backtranslated-mono) on final system quality is not experimented. Finally, participants point out one peculiarity they've noticed in the train/validation partitioning of the original MTNT dataset; validation source sentences being started with the letter ``Y" followed by alphabetically sorted sentences (test partition not effected).

\paragraph{FOKUS' submission \cite{fokus}:} This team participated in three directions: Eng$\rightarrow$Fra, Fra$\rightarrow$Eng and Jpn$\rightarrow$Eng. For the Eng$\rightarrow$Fra and Fra$\rightarrow$Eng language pairs, the submissions are \textit{unconstrained} systems, where the model was trained on the medical domain corpus provided by the WMT biomedical shared task \footnote{http://www.statmt.org/wmt19/biomedical-translation-task.html}. Despite the training data being out-of-domain, removing ``low-quality" parallel data such as ``Subtitles" as the author hypothesized helped to bring 2 to 4 BLEU points improvement over the baseline models. Their Jpn$\rightarrow$Eng submission is a \textit{constrained} system, using the same model architecture as the Eng$\rightarrow$Fra language pair. To improve robustness, they introduced synthetic noise (omitting and duplicating letters) in the training data to both source and target sentences.    

\paragraph{JHU's submission \cite{jhu}:}  
This submission participated in the Fra$\rightarrow$Eng and Jpn$\leftrightarrow$Eng tasks. The participants used data dual cross-entropy filtering for reducing the monolingual data, then back-translate these, and train their Transformer models \cite{vaswani2017attention}. They 
compared Moses tokenization+Byte Pair Encoding (BPE) \cite{sennrich-etal-2016-neural}, and sentence-piece \cite{kudo-richardson-2018-sentencepiece} (without any pre-processing) 
and found the two comparable, and that using larger sentence-piece models 
improved over smaller ones. For Jpn$\leftrightarrow$Eng (both directions) they first used both in-domain (MTNT) and out-of-domain data (other constrained), and then continued training (fine-tune) using MTNT only. They also reported many results from their hyper-parameter search (albeit without a clear recommendation). The final submission is an ensemble of 4 models.

\paragraph{NaverLabsEurope(NLE)' submission \cite{naver}:} The participants carried substantial effort to clean the CommonCrawl data, applying length filtering (length ratio threshold), language identification-based filtering, and attention based filtering.
They used the Transformer-Big architecture
for Fra$\rightarrow$Eng and Jpn$\rightarrow$Eng, and Transformer-Base for 
the Eng$\rightarrow$Jpn direction.

The participants incorporated several methods to encourage robustness (detailed ablations on the effect of each method were not provided). 
They lowercase all data. However in order to preserve casing information in the input, they propose a technique called \emph{inline casing} which adds additional casing tags (one per non-lowercased subword) in the sequence. 
%
Emojis 
were replaced with a special symbol. 
Natural noise based on manually defined noise rules was added on the source side of the training data. Lastly, MTNT monolingual data was back-translated to be used during training of the final system. 
They trained their system on all available data with special tags for each domain and for each data type e.g. real, back-translated, or noisy data. They found that adding tags is as good as fine-tuning the system, allowing for more flexibility at test time.
Their final submission with an ensemble of~6 systems for Eng$\rightarrow$Jpn and ensembles of~4 systems for the other language directions performed the best in the evaluation campaign.

\paragraph{NICT's submission \cite{nict}:} The authors used Transformer models to train their systems and employed two strategies namely: i) mixed fine-tuning and ii) multilingual models  for making the systems robust. The former helps as the in-domain data is available in a very small quantity. Using a mix of in-domain and out-domain data for fine-tuning helps overcome the problem of adjusting learning rate, applying better regularization and other complicated strategies. It is not clear how these two methods contributed towards making the models more robust. 
According to the authors, mixed fine-tuning and multilingual training (bidirectional) helped. In the error analysis, they found that their system performs poorly in translating emojis. The segmentation errors generated by KyTea resulted in further errors in the translation. 

\paragraph{NTT's submission \cite{ntt}:} The participants submitted systems for the Eng$\rightarrow$Jpn and Jpn$\rightarrow$Eng directions in the constrained setting. Their techniques include the placeholder mechanism for copying non-standard tokens (emojis, emoticons, etc), back-translation, fine-tuning on in-domain corpus, and ensemble. Especially, the placeholder mechanism provides +1.4 BLEU and +0.7 BLEU points for Jpn$\rightarrow$Eng and Eng$\rightarrow$Jpn respectively. Finetuning provides a larger improvement for Eng$\rightarrow$Jpn (+1.2 BLEU) than Jpn$\rightarrow$Eng (-0.3 BLEU). Their model is Transformer-Base configuration, where they demonstrated its capacity to noise-robustness can be further improved by the above-mentioned techniques.




\section{Summary of Methods} \label{sec:methods}

In this section, we give a common theme and summary of methods applied by the various participants. 

\paragraph{Data Cleaning}

Data cleaning played an important part in training successful MT systems in this campaign. Unlike other participants, the winning team Naver Labs \newcite{naver} and NTT \cite{ntt}
applied data cleaning techniques in order to filter noisy parallel sentences. They filtered i) identical sentences on source and target side, ii) sentences that belonged to a language other than the source and target language, iii) sentences with 
length mismatch, and iv) also applied attention-based filtering. 
Data cleaning 
gave an improvement of more than 5 BLEU points with substantial reduction in the hallucination of the model for the winning team. 

\paragraph{Placeholders}

Training and test data contained tokens (such as emoticons) which do not require translation. \newcite{ntt} and \newcite{naver} preserved 
these in a preprocessing step using special placeholders and copied them in the translation output. \newcite{ntt} 
reported a gain of up to 1.4 BLEU points by using placeholders. 

\paragraph{Data Augmentation}

Other than handling noisy data, one of the challenges related to this task was data sparsity. All the participants back-translated in-domain monolingual data and used synthetic data as part of their training pipeline. In addition, \newcite{naver} created a noisy version of all the available in-domain and out-of-domain data by randomly replacing words with their noisy variants. For training, they appended source sentences with a tag $<$noisy$>$ to distinguish them from the original data. 
\citet{cmu} used translation systems using placeholders in order to create both clean versions of the noisy in-domain datasets, as well as noisy versions of the clean out-of-domain dataset.
To get additional data, other than back-translation, the JHU team \cite{jhu} used cross-entropy based filtering to select top 1 million sentences from Gigaword, CommonCrawl and the UN corpus. Adding large filtered data gave then an improvement of +5.8 BLEU points. 


\paragraph{Domain-aware Training}

In order to differentiate different data, real from synthetic, in-domain from out-domain, several participants used additional tags. \newcite{baidu, naver} used domain tags during training to indicate data domain. 
\newcite{naver} additionally included data type tags (real or back-translated) for further categorization of the training data. 
Compared to fine-tuning, adding tags provides them additional flexibility, resulting in a generalized system, robust towards a variety of input data.

\paragraph{Fine-tuning}

Along with the noisy in-domain MTNT data, general domain data typically made available for WMT campaign was also allowed for this task. Most participants \cite{ntt,nict,cuni} trained on general domain data and fine-tuned the models towards the task.
\newcite{ntt} did not see a consistent improvement with fine-tuning. Due to the small size of the in-domain data, \newcite{nict} fine-tuned on a mix of in-domain and a subset of the out-of-domain data.

\paragraph{Ensembles}

To benefit from the different trained models and to make the performance more stable, many participants performed {\tt ensemble} over their models.  
\newcite{ntt}, \newcite{naver}, \newcite{baidu}, and \newcite{jhu} ensembled between 4 and 6 checkpoints of their model for the final submission. They observed a consistent performance improvement over using a single model. 

\section{Results} \label{sec:results}

In this section we describe quantitative results, and also perform a qualitative analysis of the results.

\begin{table*}[t]
    \centering
    \begin{tabular}{l|cc @{\hskip 3em} cc}
    \toprule
        \multirow{2}{*}{System} & \multicolumn{4}{c}{Human judgment scores (\textsc{rank})} \\
        & Eng$\rightarrow$Fra & Fra$\rightarrow$Eng & Eng$\rightarrow$Jpn & Jpn$\rightarrow$Eng \\
    \midrule
    \multicolumn{5}{@{}l}{\textit{Constrained}}\\
        Baidu+OSU & \metric{71.540090411611} \textsc{(2)} & \metric{80.573127872398	} \textsc{(3)} & -- & --\\
        CMU & -- & \metric{58.173830765072} \textsc{(6)} & -- & --  \\
        CUNI & \metric{66.330478229836} \textsc{(3)} & \metric{82.02351987} \textsc{(2)} & -- & -- \\
        FOKUS & -- & -- & -- & \metric{48.517417417417} \textsc{(5)}\\
        JHU & -- & \metric{76.33711814} \textsc{(4)} & \metric{58.540229885057} \textsc{(3)} & \metric{65.403603603604} \textsc{(3)}\\
        NaverLabs & \textbf{75.5} \textsc{(1)} & \textbf{85.3} \textsc{(1)} & \metric{63.881704980843} \textsc{(2)} & \textbf{74.1} \textsc{(1)} \\
        NTT & -- & -- & \textbf{66.5} \textsc{(1)} & \metric{71.33273273} \textsc{(2)} \\
        NICT & -- & -- & \metric{44.67241379} \textsc{(4)} & \metric{49.065765765766} \textsc{(4)}\\
    \midrule
    \multicolumn{5}{@{}l}{\textit{Unconstrained}}\\
    FOKUS & \metric{52.45324768} \textsc{(4)} & \metric{62.59529603} \textsc{(5)} & -- & --\\
        
    \bottomrule
    \end{tabular}
    \caption{Average human judgments over all submitted systems (the higher the better). The systems' rank for each translation direction is shown in parentheses. The best system is \textbf{highlighted}.}
    \label{tab:results_human}
\end{table*}
\begin{table*}[t]
    \centering
    \begin{tabular}{l|cc @{\hskip 3em} cc}
    \toprule
        \multirow{2}{*}{System} & \multicolumn{4}{c}{BLEU \textsc{(rank)}} \\
        & Eng$\rightarrow$Fra & Fra$\rightarrow$Eng & Eng$\rightarrow$Jpn & Jpn$\rightarrow$Eng \\
    \midrule
        Baseline & $22.1$ & $25.6$ & $8.4$ & $5.8$ \\
    \midrule
    \multicolumn{5}{@{}l}{\textit{Constrained}}\\
        Baidu+OSU & $36.39$ \textsc{(3)} & $43.59$ \textsc{(3)} & -- & --\\
        CMU & -- & $32.25$ \textsc{(5)} & -- & --  \\
        CUNI & $38.49$ \textsc{(2)} & $44.83$ \textsc{(2)} & -- & -- \\
        FOKUS & -- & -- & -- & $6.42$ \textsc{(5)} \\        JHU & -- & $40.24$ \textsc{(4)} & $14.67$ \textsc{(3)} & $12.01$ \textsc{(3)} \\
        NaverLabs & $\mathbf{41.39}$ \textsc{(1)} & $\mathbf{47.93}$ \textsc{(1)} & $\mathbf{17.73}$ \textsc{(1)} & $\mathbf{16.41}$ \textsc{(1)}\\
        NTT & -- & -- & $16.86$ \textsc{(2)} & $14.82$ \textsc{(2)} \\
        NICT & -- & -- & $11.09$ \textsc{(4)} & $7.56$ \textsc{(4)} \\
    \midrule
    \multicolumn{5}{@{}l}{\textit{Unconstrained}}\\
        FOKUS & $24.22$ \textsc{(4)} & $29.94$ \textsc{(6)} & -- & -- \\
    \bottomrule
    \end{tabular}
    \caption{Automatic evaluation (BLEU, cased) over all submitted systems, with the system's rank in parentheses. The best system is \textbf{highlighted}.}
    \label{tab:results_bleu}
\end{table*}

\subsection{Quantitative Results}

The quantitative analysis of the submitted systems yields fairly consistent results. On automatic evaluation (BLEU) the best system across all translation directions is the NaverLabsEurope(NLE) one. The same system received also the highest human judgment scores, with the exception of the Eng$\rightarrow$Jpn task, where the NTT system was ranked higher.
Overall, the correlation between human judgments and BLEU is very high. For Eng$\rightarrow$Fra, the Pearson's correlation coefficient is~$0.94$, while for the other three tasks it is over~$0.97$.

\paragraph{Human Evaluation}
The results of human evaluation following the evaluation protocol described in Section~\ref{sec:protocol} are outlined in Table~\ref{tab:results_human}.

\paragraph{Automatic Evaluation}
The automatic evaluation (BLEU) results of the Shared Task are summarized in Table~\ref{tab:results_bleu}. 

\subsection{Qualitative Analysis}

In order to discover salient differences between the methods, we performed analysis using \texttt{compare-mt} \cite{neubig19comparemt}, and present a few of the salient findings below.

\paragraph{Stronger Submissions were Stronger at Everything:}
The submissions to the track achieved a wide range of BLEU and human evaluation scores.
In our analysis we found that the systems at the higher end of the spectrum with regards to BLEU also tended to be the best by most other measures (human evaluation, word F-measure by various frequency buckets, sentence-level scores, etc.).
Because of this, we limit our remaining analysis to the top three systems in the Fra$\rightarrow$Eng and Eng$\rightarrow$Fra tracks, and the top two systems in the Eng$\rightarrow$Jpn and Jpn$\rightarrow$Eng tracks.

\begin{figure}[h]
  \centering
  \includegraphics[width=\linewidth]{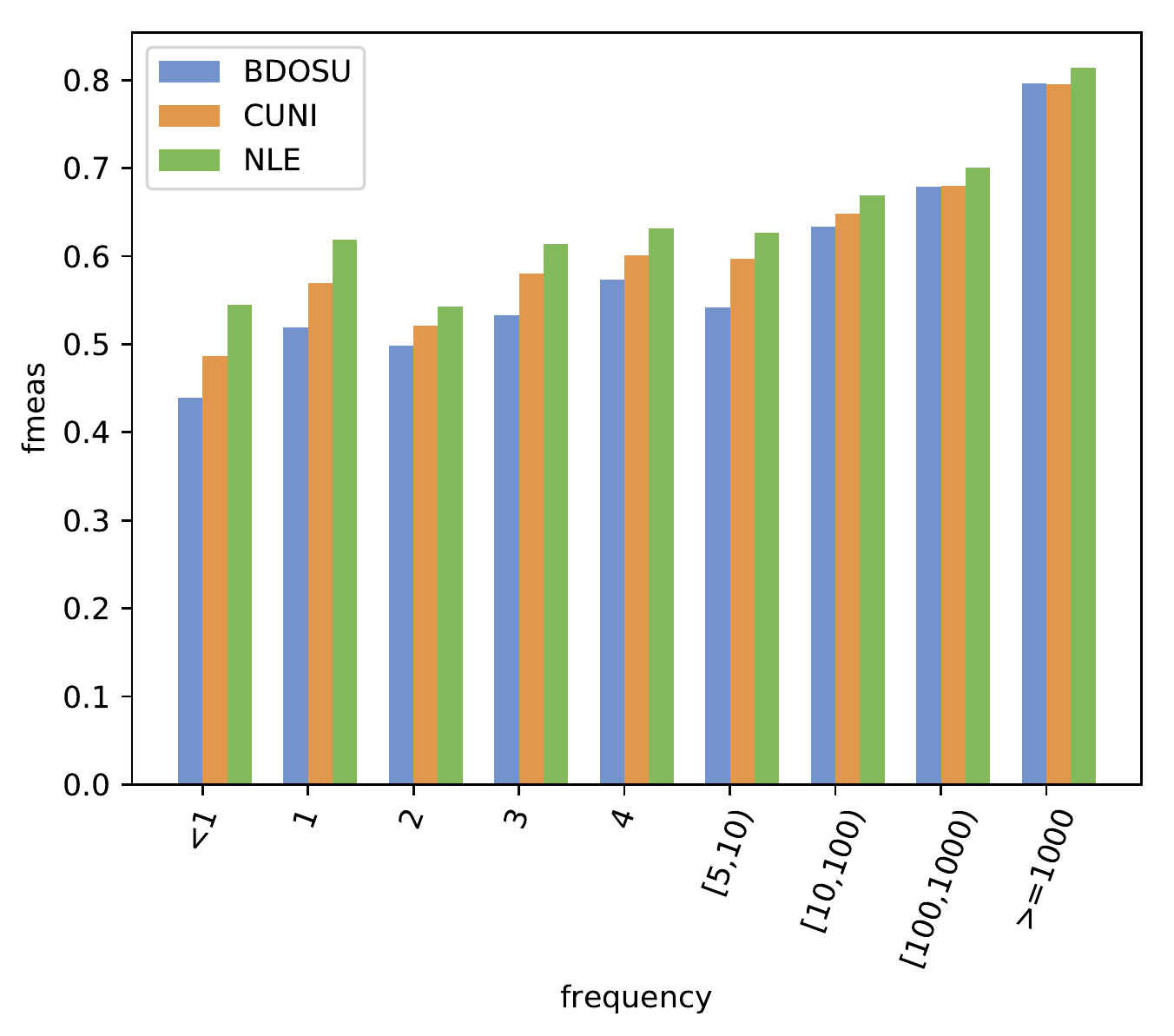}
  \caption{Word F-measure by frequency in the MTNT training data for Fra-Eng.}
  \label{fig:fr-en-freq-accuracy}
\end{figure}

\paragraph{Generalization to Words not in Adaptation Data is Essential:}
The MTNT corpus provides a small amount of training data that can be used to adapt systems to the task of translating social media.
One large distinguishing factor between the best-performing system by Naver Labs Europe (NLE) and the second- or third-place systems was performance on words that were \emph{not} included in this training data that nonetheless appeared in the test set.
We show the example of word-level F-measure bucketed by frequency of the words in the MTNT test set for Fra$\rightarrow$Eng in Figure \ref{fig:fr-en-freq-accuracy}.
From this figure we can see that the NLE system does a bit better in all frequency categories, but the difference is particularly stark for words that appear only once or not at all in the MTNT training set.

\begin{figure}[h]
  \centering
  \includegraphics[width=\linewidth]{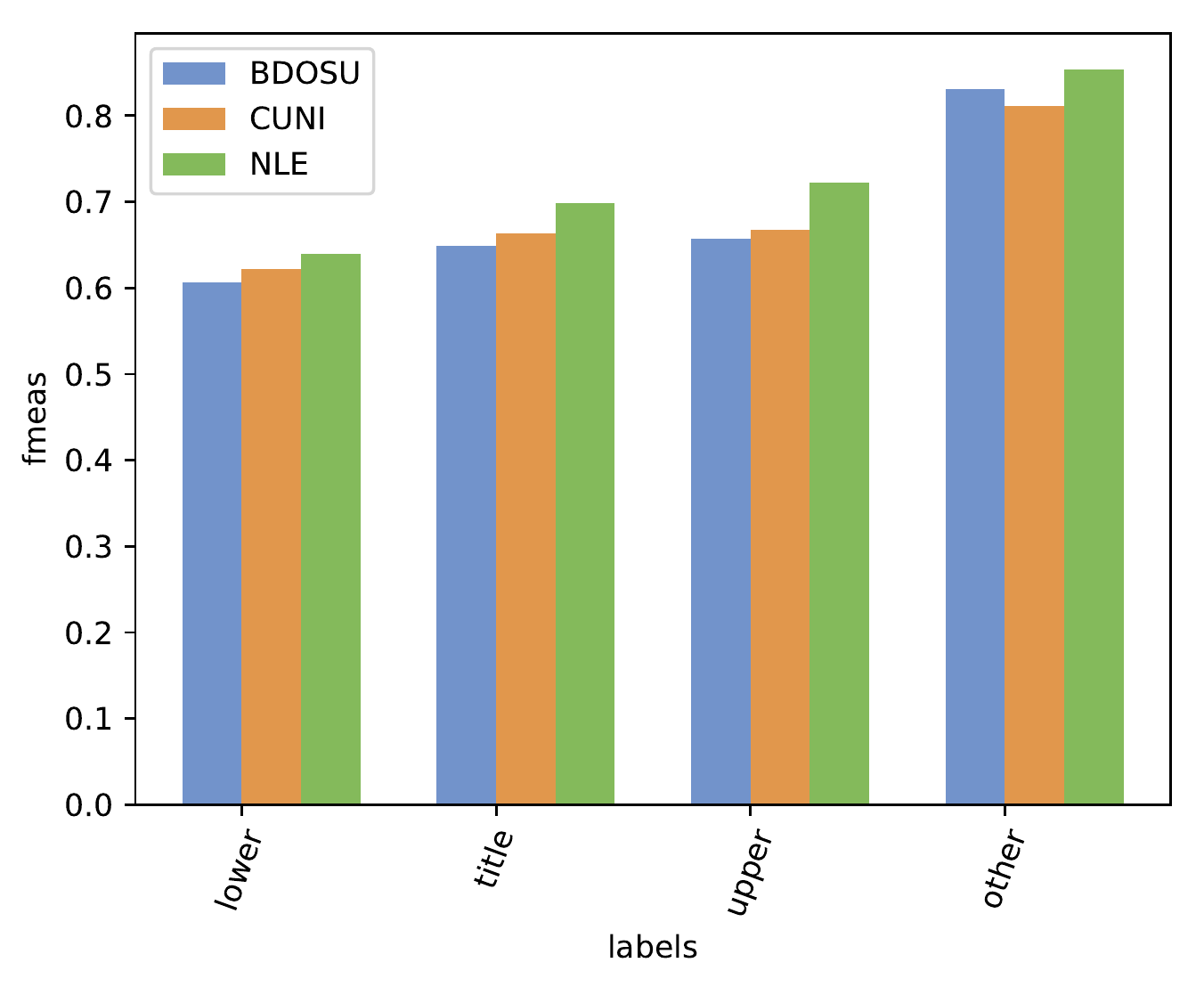}
  \caption{Word F-measure by casing of the words in the target: all lower-case, title case, all upper-case, or other.}
  \label{fig:en-fr-casing-accuracy}
\end{figure}

\begin{table*}[t]
  \centering
  \begin{tabular}{c | p{10cm}c}
  \toprule 
 & Output & BLEU+1 \\ \midrule 
Ref & From Sri Lanka , to Russia , to the United States , to Japan I mean it 's a market THAT GOES EVERYWHERE . &  \\
CUNI & from sri lanka , to russia , to the united states , to japon I mean it 's a market QUI VA PARTOUT . & 33.0 \\
NLE & From Sri Lanka , to Russia , to the United States , to Japan I mean it 's a market THAT GOES EVERYWHERE . & 100 \\
\bottomrule 
  \end{tabular}
  \caption{An example of handling of casing in two Fra$\rightarrow$Eng systems}
  \label{tab:fr-en-casing-example}
\end{table*}

\paragraph{Proper Handling of Casing is Important:}
One other innovation performed by the NLE team was lowercasing of words and separate prediction of casing information.
This modeling decision apparently resulted in significantly better results particularly on words that were written in all upper-case, as demonstrated in the results of word F-measure by casing in the target language, demonstrated for Eng$\rightarrow$Fra in Figure~\ref{fig:en-fr-casing-accuracy}.
In addition, we show an example for Fra$\rightarrow$Eng in Table \ref{tab:fr-en-casing-example}, where the NLE system translates upper-case characters perfectly, but the CUNI system struggles.

\begin{table}[t]
  \centering
  \begin{tabular}{c | p{4.5cm}c}
  \toprule 
 & Output & BLEU+1 \\ 
 \midrule 
Ref & Kawaii \emoji{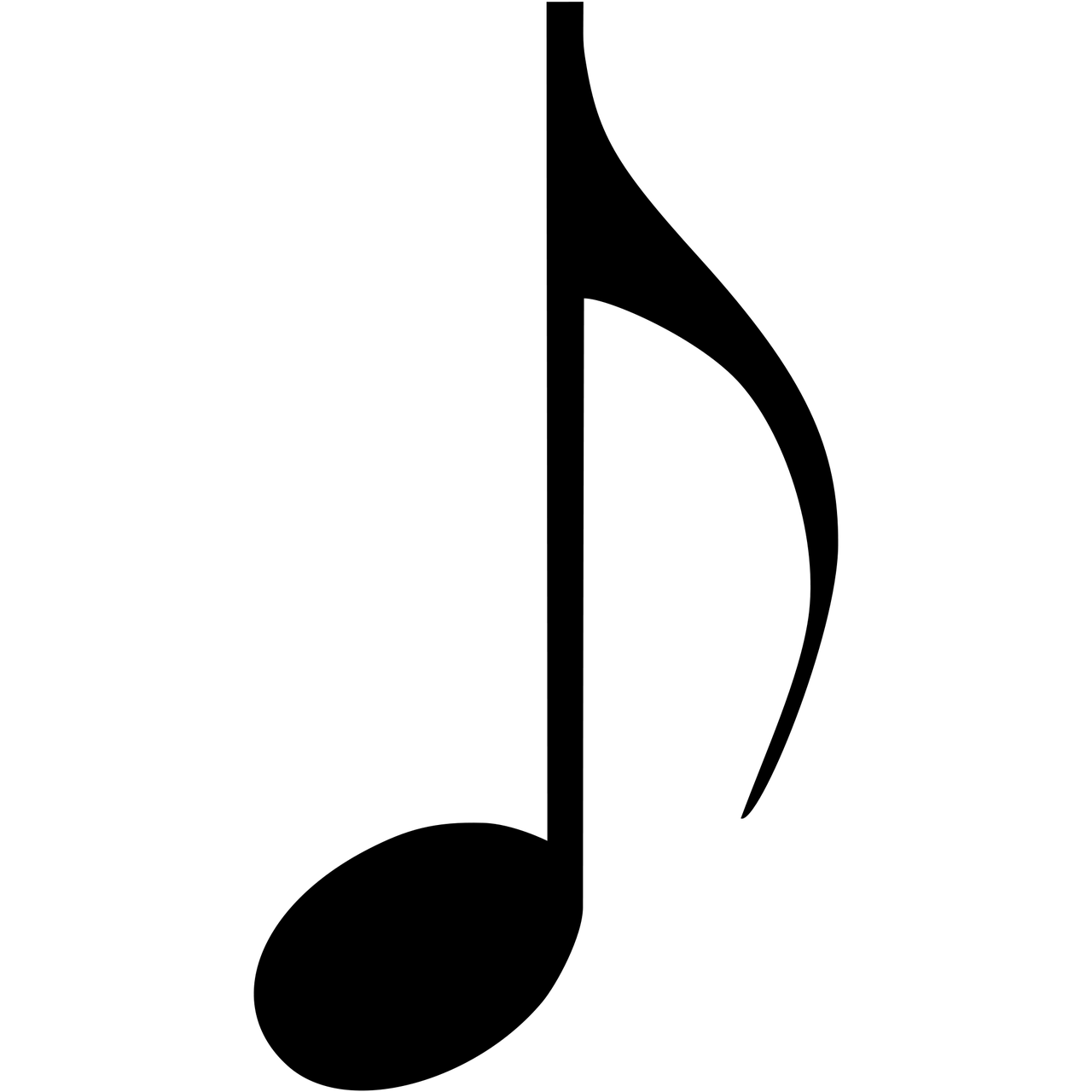} \begin{CJK*}{UTF8}{min} (*・ω・人)\end{CJK*} &  \\
NTT & Cute \begin{CJK*}{UTF8}{min}(*・ω・人)\end{CJK*} & 76.0 \\
NLE & It 's cute . & 0.0 \\ 
\midrule 
Ref & \emoji{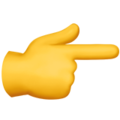} \emoji{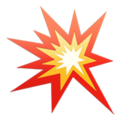} \emoji{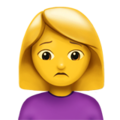} \emoji{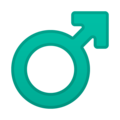} &  \\
NTT &  & 0.0 \\
NLE & \emoji{white-right-pointing-backhand} \emoji{collision} \emoji{person-frowning} \emoji{male-sign} & 100 \\
\bottomrule 
\end{tabular}
  \caption{Examples of translation results on special characters.}
  \label{tab:ja-en-emojis}
\end{table}

\paragraph{Special Handling of Special Characters is Beneficial:}
Special characters such as Emojis or symbols were difficult for some systems.
Interestingly, even among the top systems, some systems were better at handling different varieties of these characters than others.
As an example, in Jpn$\rightarrow$Eng, the NTT system performed better on Japanese-style smileys written with standard characters, while the NLE system performed better on Unicode-standard Emojis, as shown in Table \ref{tab:ja-en-emojis}.

\begin{table*}[t]
  \centering
  \begin{tabular}{c | p{10cm}c}
  \toprule 
 & Output & BLEU+1 \\ 
 \midrule 
Ref & * * ] ( \# mm-e9 ) [ * * Because there 's now protection * * ] ( \# mm-e4 ) &  \\
NTT & * * * * ) ( \# m-e9 ) [ * * * * * * * * * * * * * * * * * -e4 because there is more protection . ) & 14.3 \\
NLE & * * * ( \# mm-e9 ) [ * * Because there is already protection * * ] ( \# mm-e4 ) & 72.0 \\
\bottomrule 
  \end{tabular}
  \caption{An example of translation results on as sentence with an unusual number of special symbols.}
  \label{tab:ja-en-nonstandard}
\end{table*}

\paragraph{Non-standard Sentence Structure can be Difficult:}
Some systems also found sentences with unusual structures, including brackets or other types of punctuation interspersed with actual text, particularly difficult.
For example, Table \ref{tab:ja-en-nonstandard} shows an example of Jpn$\rightarrow$Eng sentences where the NTT system had trouble generating the appropriate number of symbols in the appropriate places, while the NLE system was more robust in this regard.

\begin{table}[t]
  \centering
  \begin{tabular}{ll | rr}
  \toprule 
   & n-gram      & NTT & NLE \\ 
   \midrule 
1. & \begin{CJK*}{UTF8}{min}て い な い \end{CJK*} & 5 & 0 \\
2. & \begin{CJK*}{UTF8}{min}だ けど    \end{CJK*} & 4 & 0 \\
3. & \begin{CJK*}{UTF8}{min}多く の    \end{CJK*} & 4 & 0 \\
4. & \begin{CJK*}{UTF8}{min}ね         \end{CJK*} & 3 & 0 \\
5. & \begin{CJK*}{UTF8}{min}だ けど 、 \end{CJK*} & 3 & 0 \\
\bottomrule 
\end{tabular}
  \caption{Examples of n-grams where one the NTT Eng$\rightarrow$Jpn system was more accurate than the NLE system}
  \label{tab:en-ja-colloquial}
\end{table}

\paragraph{Colloquial Expressions are Key:}
There was also a marked difference among the top systems in their ability to produce the more informal register reflected in the MTNT test data.
We show an example in Table \ref{tab:en-ja-colloquial} of $n$-grams that the NTT system was better at producing than the NLE system.
All of these are relatively colloquial ways of expressing common function word phrases (1.~``is not doing'', 2.~``but'', 3.~``lots'', 4.~``right?'', 5.~``but,'') that can also be expressed with more formal expressions.
Clearly the NTT system is producing a slightly less formal register than the NLE system, although a manual examination of the outputs found that even the NTT system was still commonly producing register that was more formal than is commonly found on social media.
This may be attributed to the fact that the NTT system performed fine-tuning on the MTNT data, moving it towards a more appropriately colloquial register.

\section{Conclusions}
As a new WMT shared task, this year we focused on building MT systems which are robust to input variations commonly observed in informal language, social media text etc. 

From a methodological perspective, the ``constrained" setup of the task encouraged participants to leverage both out-of-domain parallel data and in-domain monolingual data to improve performance. Some techniques were utilized by multiple participants and proved their effectiveness in boosting MT models' robustness to noisy input and domain mismatch, including data cleaning, domain-aware training, data augmentation (including backtranslation and copying place-holder tags), finetuning, etc. 

In terms of evaluation, we found an automatic metric (BLEU) to be roughly consistent with human judgment.
Qualitative analysis found that strong baseline systems were important, but on top of this additional methods specifically aimed at trying to handle various types of noise found in social media text were effective and necessary to further improve within the upper echelons of systems submitted to the shared task.

There are several directions to be explored in the future editions of the task. First, it can exhibit a separate track for ``probing'' models' robustness so as to understand current models' weaknesses. Second, it could further disentangle improvements for different challenges, e.g., due to noise in training data or due to distribution shift at test time. Controlling the kind of noise introduced, e.g. natural vs.\ artificial, may be useful in this regard. 

\section*{Acknowledgements}
We thank Facebook for funding the human evaluation and blind test set creation. 

\bibliographystyle{acl_natbib}
\bibliography{acl2019}

\begin{thebibliography}{37}
\expandafter\ifx\csname natexlab\endcsname\relax\def\natexlab#1{#1}\fi

\bibitem[{Anastasopoulos et~al.(2019)Anastasopoulos, Lui, Nguyen, and
  Chiang}]{anastasopoulos+etal:naacl2019}
Antonios Anastasopoulos, Alison Lui, Toan~Q. Nguyen, and David Chiang. 2019.
\newblock Neural machine translation of text from non-native speakers.
\newblock In \emph{Proc. NAACL HLT}.

\bibitem[{Belinkov and Bisk(2018)}]{belinkov2017synthetic}
Yonatan Belinkov and Yonatan Bisk. 2018.
\newblock Synthetic and natural noise both break neural machine translation.
\newblock In \emph{International Conference on Learning Representations
  (ICLR)}.

\bibitem[{Belinkov and Glass(2019)}]{belinkov2019analysis}
Yonatan Belinkov and James Glass. 2019.
\newblock \href {https://doi.org/10.1162/tacl\_a\_00254} {Analysis methods in
  neural language processing: A survey}.
\newblock \emph{Transactions of the Association for Computational Linguistics
  (TACL)}, 7:49--72.

\bibitem[{B{\'e}rard et~al.(2019)B{\'e}rard, Calapodescu, and Roux}]{naver}
Alexandre B{\'e}rard, Ioan Calapodescu, and Claude Roux. 2019.
\newblock {Naver Labs Europe’s Systems for the WMT19 Machine Translation
  Robustness Task}.
\newblock In \emph{Proceedings of the 2019 Shared task on Machine Translation
  Robustness, Conference on Machine Translation (WMT)}.

\bibitem[{Cettolo et~al.(2012)Cettolo, Girardi, and
  Federico}]{cettoloEtAl:EAMT2012}
Mauro Cettolo, Christian Girardi, and Marcello Federico. 2012.
\newblock Wit$^3$: Web inventory of transcribed and translated talks.
\newblock In \emph{Proceedings of the 16$^{th}$ Conference of the European
  Association for Machine Translation (EAMT)}, pages 261--268.

\bibitem[{Cheng et~al.(2018{\natexlab{a}})Cheng, Yi, Zhang, Chen, and
  Hsieh}]{Cheng2018Seq2SickET}
Minhao Cheng, Jinfeng Yi, Huan Zhang, Pin-Yu Chen, and Cho-Jui Hsieh.
  2018{\natexlab{a}}.
\newblock Seq2sick: Evaluating the robustness of sequence-to-sequence models
  with adversarial examples.
\newblock \emph{CoRR}, abs/1803.01128.

\bibitem[{Cheng et~al.(2019)Cheng, Jiang, and
  Macherey}]{Cheng-etal-2019-doubly}
Yong Cheng, Lu~Jiang, and Wolfgang Macherey. 2019.
\newblock Robust neural machine translation with doubly adversarial inputs.
\newblock In \emph{ACL}. Association for Computational Linguistics.

\bibitem[{Cheng et~al.(2018{\natexlab{b}})Cheng, Tu, Meng, Zhai, and
  Liu}]{cheng-tu-2018-towards}
Yong Cheng, Zhaopeng Tu, Fandong Meng, Junjie Zhai, and Yang Liu.
  2018{\natexlab{b}}.
\newblock \href {http://arxiv.org/abs/1805.06130} {Towards robust neural
  machine translation}.
\newblock \emph{CoRR}, abs/1805.06130.

\bibitem[{Dabre and Sumita(2019)}]{nict}
Raj Dabre and Eiichiro Sumita. 2019.
\newblock {NICT’s Supervised MT Systems for the Translation Robustness Task
  in WMT19}.
\newblock In \emph{Proceedings of the 2019 Shared task on Machine Translation
  Robustness, Conference on Machine Translation (WMT)}.

\bibitem[{Durrani et~al.(2019)Durrani, Dalvi, Sajjad, Belinkov, and
  Nakov}]{durrani-etal-2019}
Nadir Durrani, Fahim Dalvi, Hassan Sajjad, Yonatan Belinkov, and Preslav Nakov.
  2019.
\newblock \href {https://www.aclweb.org/anthology/N19-1154} {One size does not
  fit all: Comparing {NMT} representations of different granularities}.
\newblock In \emph{Proceedings of the 2019 Conference of the North {A}merican
  Chapter of the Association for Computational Linguistics: Human Language
  Technologies, Volume 1 (Long and Short Papers)}, pages 1504--1516,
  Minneapolis, Minnesota. Association for Computational Linguistics.

\bibitem[{Ebrahimi et~al.(2018)Ebrahimi, Lowd, and
  Dou}]{ebrahimi-etal-2018-adversarial}
Javid Ebrahimi, Daniel Lowd, and Dejing Dou. 2018.
\newblock On adversarial examples for character-level neural machine
  translation.
\newblock In \emph{Proceedings of the 27th International Conference on
  Computational Linguistics}, Santa Fe, New Mexico, USA. Association for
  Computational Linguistics.

\bibitem[{Grozea(2019)}]{fokus}
Cristian Grozea. 2019.
\newblock {The submission of FOKUS to the WMT 19 robustness task}.
\newblock In \emph{Proceedings of the 2019 Shared task on Machine Translation
  Robustness, Conference on Machine Translation (WMT)}.

\bibitem[{Hassan et~al.(2018)Hassan, Aue, Chen, Chowdhary, Clark, Federmann,
  Huang, Junczys-Dowmunt, Lewis, Li et~al.}]{hassan2018achieving}
Hany Hassan, Anthony Aue, Chang Chen, Vishal Chowdhary, Jonathan Clark,
  Christian Federmann, Xuedong Huang, Marcin Junczys-Dowmunt, William Lewis,
  Mu~Li, et~al. 2018.
\newblock Achieving human parity on automatic chinese to english news
  translation.
\newblock \emph{arXiv preprint arXiv:1803.05567}.

\bibitem[{Heigold et~al.(2017)Heigold, Neumann, and van
  Genabith}]{heigold2017robust}
Georg Heigold, G{\"u}nter Neumann, and Josef van Genabith. 2017.
\newblock How robust are character-based word embeddings in tagging and mt
  against wrod scramlbing or randdm nouse?
\newblock \emph{arXiv preprint arXiv:1704.04441}.

\bibitem[{Helcl et~al.(2019)Helcl, Libovick{\'y}, and Popel}]{cuni}
Jind{\v r}ich Helcl, Jind{\v r}ich Libovick{\'y}, and Martin Popel. 2019.
\newblock {CUNI System for the WMT19 Robustness Task}.
\newblock In \emph{Proceedings of the 2019 Shared task on Machine Translation
  Robustness, Conference on Machine Translation (WMT)}.

\bibitem[{Karpukhin et~al.(2019)Karpukhin, Levy, Eisenstein, and
  Ghazvininejad}]{DBLP:journals/corr/abs-1902-01509}
Vladimir Karpukhin, Omer Levy, Jacob Eisenstein, and Marjan Ghazvininejad.
  2019.
\newblock \href {http://arxiv.org/abs/1902.01509} {Training on synthetic noise
  improves robustness to natural noise in machine translation}.
\newblock \emph{CoRR}, abs/1902.01509.

\bibitem[{Khayrallah and Koehn(2018)}]{khayrallah-koehn-2018-impact}
Huda Khayrallah and Philipp Koehn. 2018.
\newblock \href {https://www.aclweb.org/anthology/W18-2709} {On the impact of
  various types of noise on neural machine translation}.
\newblock In \emph{Proceedings of the 2nd Workshop on Neural Machine
  Translation and Generation}, pages 74--83, Melbourne, Australia. Association
  for Computational Linguistics.

\bibitem[{Klein et~al.(2017)Klein, Kim, Deng, Senellart, and Rush}]{opennmt}
Guillaume Klein, Yoon Kim, Yuntian Deng, Jean Senellart, and Alexander~M. Rush.
  2017.
\newblock \href {https://doi.org/10.18653/v1/P17-4012} {Open{NMT}: Open-source
  toolkit for neural machine translation}.
\newblock In \emph{Proc. ACL}.

\bibitem[{Koehn and Knowles(2017)}]{koehn2017six}
Philipp Koehn and Rebecca Knowles. 2017.
\newblock Six challenges for neural machine translation.
\newblock \emph{arXiv preprint arXiv:1706.03872}.

\bibitem[{Kudo and Richardson(2018)}]{kudo-richardson-2018-sentencepiece}
Taku Kudo and John Richardson. 2018.
\newblock \href {https://www.aclweb.org/anthology/D18-2012} {{S}entence{P}iece:
  A simple and language independent subword tokenizer and detokenizer for
  neural text processing}.
\newblock In \emph{Proceedings of the 2018 Conference on Empirical Methods in
  Natural Language Processing: System Demonstrations}, pages 66--71, Brussels,
  Belgium. Association for Computational Linguistics.

\bibitem[{Lee et~al.(2018)Lee, Firat, Agarwal, Fannjiang, and
  Sussillo}]{Lee2018HallucinationsIN}
Katherine Lee, Orhan Firat, Ashish Agarwal, Clara Fannjiang, and David
  Sussillo. 2018.
\newblock Hallucinations in neural machine translation.
\newblock In \emph{Interpretability and Robustness in Audio, Speech, and
  Language Workshop Conference on Neural Information Processing Systems}.

\bibitem[{Michel et~al.(2019)Michel, Li, Neubig, and
  Pino}]{michel2019evaluation}
Paul Michel, Xian Li, Graham Neubig, and Juan~Miguel Pino. 2019.
\newblock On evaluation of adversarial perturbations for sequence-to-sequence
  models.
\newblock In \emph{Proc. NAACL HLT}.

\bibitem[{Michel and Neubig(2018)}]{michel2018mtnt}
Paul Michel and Graham Neubig. 2018.
\newblock {MTNT}: A testbed for {M}achine {T}ranslation of {N}oisy {T}ext.
\newblock In \emph{Proceedings of the 2018 Conference on Empirical Methods in
  Natural Language Processing (EMNLP)}.

\bibitem[{Murakami et~al.(2019)Murakami, Morishita, Hirao, and Nagata}]{ntt}
Soichiro Murakami, Makoto Morishita, Tsutomu Hirao, and Masaaki Nagata. 2019.
\newblock {NTT’s Machine Translation Systems for WMT19 Robustness Task}.
\newblock In \emph{Proceedings of the 2019 Shared task on Machine Translation
  Robustness, Conference on Machine Translation (WMT)}.

\bibitem[{Neubig(2011)}]{neubig11kftt}
Graham Neubig. 2011.
\newblock The {Kyoto} free translation task.
\newblock http://www.phontron.com/kftt.

\bibitem[{Neubig et~al.(2019)Neubig, Dou, Hu, Michel, Pruthi, and
  Wang}]{neubig19comparemt}
Graham Neubig, Zi-Yi Dou, Junjie Hu, Paul Michel, Danish Pruthi, and Xinyi
  Wang. 2019.
\newblock \href {https://www.aclweb.org/anthology/N19-4007} {compare-mt: A tool
  for holistic comparison of language generation systems}.
\newblock In \emph{Proceedings of the 2019 Conference of the North {A}merican
  Chapter of the Association for Computational Linguistics (Demonstrations)},
  pages 35--41, Minneapolis, Minnesota. Association for Computational
  Linguistics.

\bibitem[{Papineni et~al.(2002)Papineni, Roukos, Ward, and
  Zhu}]{papineni-etal-2002-bleu}
Kishore Papineni, Salim Roukos, Todd Ward, and Wei-Jing Zhu. 2002.
\newblock \href {https://doi.org/10.3115/1073083.1073135} {{B}leu: a method for
  automatic evaluation of machine translation}.
\newblock In \emph{Proceedings of 40th Annual Meeting of the Association for
  Computational Linguistics}, pages 311--318, Philadelphia, Pennsylvania, USA.
  Association for Computational Linguistics.

\bibitem[{Post(2018)}]{post-2018-call}
Matt Post. 2018.
\newblock \href {https://www.aclweb.org/anthology/W18-6319} {A call for clarity
  in reporting {BLEU} scores}.
\newblock In \emph{Proceedings of the Third Conference on Machine Translation:
  Research Papers}, pages 186--191, Belgium, Brussels. Association for
  Computational Linguistics.

\bibitem[{Post and Duh(2019)}]{jhu}
Matt Post and Kevin Duh. 2019.
\newblock {JHU 2019 Robustness Task System Description}.
\newblock In \emph{Proceedings of the 2019 Shared task on Machine Translation
  Robustness, Conference on Machine Translation (WMT)}.

\bibitem[{{Pryzant} et~al.(){Pryzant}, {Chung}, {Jurafsky}, and
  {Britz}}]{pryzant_jesc_2017}
R.~{Pryzant}, Y.~{Chung}, D.~{Jurafsky}, and D.~{Britz}.
\newblock \href {http://arxiv.org/abs/1710.10639} {Jesc: Japanese-english
  subtitle corpus}.
\newblock \emph{ArXiv e-prints}.

\bibitem[{Sennrich et~al.(2016)Sennrich, Haddow, and
  Birch}]{sennrich-etal-2016-neural}
Rico Sennrich, Barry Haddow, and Alexandra Birch. 2016.
\newblock \href {https://doi.org/10.18653/v1/P16-1162} {Neural machine
  translation of rare words with subword units}.
\newblock In \emph{Proceedings of the 54th Annual Meeting of the Association
  for Computational Linguistics (Volume 1: Long Papers)}, pages 1715--1725,
  Berlin, Germany. Association for Computational Linguistics.

\bibitem[{Szegedy et~al.(2013)Szegedy, Zaremba, Sutskever, Bruna, Erhan,
  Goodfellow, and Fergus}]{szegedy2013intriguing}
Christian Szegedy, Wojciech Zaremba, Ilya Sutskever, Joan Bruna, Dumitru Erhan,
  Ian Goodfellow, and Rob Fergus. 2013.
\newblock Intriguing properties of neural networks.
\newblock \emph{arXiv preprint arXiv:1312.6199}.

\bibitem[{Vaibhav et~al.(2019)Vaibhav, Singh, Stewart, and
  Neubig}]{Vaibhav-etal-2019-improving}
Vaibhav Vaibhav, Sumeet Singh, Craig Stewart, and Graham Neubig. 2019.
\newblock \href {https://www.aclweb.org/anthology/N19-1190} {Improving
  robustness of machine translation with synthetic noise}.
\newblock In \emph{Proceedings of the 2019 Conference of the North {A}merican
  Chapter of the Association for Computational Linguistics: Human Language
  Technologies, Volume 1 (Long and Short Papers)}, pages 1916--1920,
  Minneapolis, Minnesota. Association for Computational Linguistics.

\bibitem[{Vaswani et~al.(2017)Vaswani, Shazeer, Parmar, Uszkoreit, Jones,
  Gomez, Kaiser, and Polosukhin}]{vaswani2017attention}
Ashish Vaswani, Noam Shazeer, Niki Parmar, Jakob Uszkoreit, Llion Jones,
  Aidan~N Gomez, {\L}ukasz Kaiser, and Illia Polosukhin. 2017.
\newblock Attention is all you need.
\newblock In \emph{Advances in neural information processing systems}, pages
  5998--6008.

\bibitem[{Zhao et~al.(2018)Zhao, Dua, and Singh}]{Zhao2018generating}
Zhengli Zhao, Dheeru Dua, and Sameer Singh. 2018.
\newblock \href {https://openreview.net/forum?id=H1BLjgZCb} {Generating natural
  adversarial examples}.
\newblock In \emph{International Conference on Learning Representations}.

\bibitem[{Zheng et~al.(2019)Zheng, Liu, Ma, Zheng, and Huang}]{baidu}
Renjie Zheng, Hairong Liu, Mingbo Ma, Baigong Zheng, and Liang Huang. 2019.
\newblock {Robust Machine Translation with Domain Sensitive Pseudo-Sources:
  Baidu-OSU WMT19 MT Robustness Shared Task System Report}.
\newblock In \emph{Proceedings of the 2019 Shared task on Machine Translation
  Robustness, Conference on Machine Translation (WMT)}.

\bibitem[{Zhou et~al.(2019)Zhou, Zeng, Zhou, Anastasopoulos, and Neubig}]{cmu}
Shuyan Zhou, Xiangkai Zeng, Yingqi Zhou, Antonios Anastasopoulos, and Graham
  Neubig. 2019.
\newblock {Improving Robustness of Neural Machine Translation with Multi-task
  Learning}.
\newblock In \emph{Proceedings of the 2019 Shared task on Machine Translation
  Robustness, Conference on Machine Translation (WMT)}.

\end{thebibliography}

\end{document}